\documentclass{article}
\usepackage[pagebackref=true,breaklinks=true,letterpaper=true,colorlinks,bookmarks=false]{hyperref}

     \PassOptionsToPackage{numbers}{natbib}
\usepackage[final]{neurips_2023}
\usepackage[utf8]{inputenc} % allow utf-8 input
\usepackage[T1]{fontenc}    % use 8-bit T1 fonts
\usepackage{hyperref}       % hyperlinks
\usepackage{url}            % simple URL typesetting
\usepackage{booktabs}       % professional-quality tables
\usepackage{amsfonts}       % blackboard math symbols
\usepackage{nicefrac}       % compact symbols for 1/2, etc.
\usepackage{microtype}      % microtypography
\usepackage{graphicx}% colors
\usepackage[dvipsnames]{xcolor}
\usepackage{wrapfig}

\usepackage{stmaryrd}
\usepackage{amsmath} 
\usepackage{xspace}

\usepackage{algorithm}
\usepackage{algpseudocode}

\makeatletter
\DeclareRobustCommand\onedot{\futurelet\@let@token\@onedot}
\def\@onedot{\ifx\@let@token.\else.\null\fi\xspace}

\def\eg{\emph{e.g}\onedot} 
\def\ie{\emph{i.e}\onedot}

\def\wrt{w.r.t\onedot} 
\def\etal{\emph{et al}\onedot}
\makeatother

\title{Robustifying Generalizable Implicit Shape Networks with a Tunable Non-Parametric Model}

\author{%
  Amine Ouasfi \quad \quad Adnane Boukhayma\\
  Inria, Univ. Rennes, CNRS, IRISA, M2S, France
}

\begin{document}

\maketitle

\begin{abstract}
Feedforward generalizable models for implicit shape reconstruction from unoriented point cloud present multiple advantages, including high performance and inference speed. However, they still suffer from generalization issues, ranging from underfitting the input point cloud, to misrepresenting samples outside of the training data distribution, or with toplogies unseen at training.  We propose here an efficient mechanism to remedy some of these limitations at test time. We combine the inter-shape data prior of the network with an intra-shape regularization prior of a Nystr\"{o}m Kernel Ridge Regression, that we further adapt by fitting its hyperprameters to the current shape. The resulting shape function defined in a shape specific Reproducing Kernel Hilbert Space benefits from desirable stability and efficiency properties and grants a shape adaptive expressiveness-robustness trade-off. We demonstrate the improvement obtained through our method  with respect to baselines and the state-of-the-art using synthetic and real data.

%Given a pretrained convolutional deep network for shape reconstruction, we define our final implicit shape function as a kernel ridge regression in the network's feature space, and we tune its hyperparameters using gradient descent. The regression is guided by the input point cloud, in addition to pseudo-labels transferred from the network. We demonstrate the improvement obtained through our method  with respect to the baselines and the state-of-the-art using synthetic and real data.
\end{abstract}

\section{Introduction}

Enabling machines to understand and navigate 3D is a major challenge. Instances of this ability include downstream tasks in computer vision and graphics, such as shape reconstruction from noisy, incomplete and relatively sparse point clouds. Building full shapes from point clouds is all the more an important problem in account of the ubiquity of this light, albeit incomplete, 3D shape representation, whether it is acquired from the increasingly democratized consumer grade and industrial depth sensors or \eg as obtained from photogrammetry (\eg Structure From Motion, Multi-View Stereo \cite{schoenberger2016sfm,schoenberger2016mvs}). Whilst Classical optimization based approaches such as Poisson Reconstruction \cite{kazhdan2013screened} and Moving least squares \cite{guennebaud2007algebraic} can deliver mostly good reconstructions for all intents and purposes, they require still dense clean point sets with reliable normal estimations. More recently, deep learning based alternatives have been shown to offer faster and more robust predictions especially for noisy and sparse inputs, without requiring normal information.

Within this class of methods, state-of-the-art feed forward (test time optimization-free) generalizable models are based on implicit neural shape representations, where explicit meshes can be extracted at test time using Marching Cubes \cite{lorensen1987marching}. While they offer many desirable properties, such as all-round good performance and fast inference (within tens of milliseconds on a standard GPU), they still suffer from limitations, some of which we come across upon examining the state-of-the-art method \cite{boulch2022poco} for instance:\\
\noindent \textbf{Limited OOD generalization}\quad As these models are typically trained with full 3D supervision, training commonly uses synthetic data such as the ShapeNet \cite{shapenet} dataset, large real 3D supervision being famously hard to come by. This results in a performance drop for real test data outside the training distribution. Comparing Tables \ref{tab:sncd} and \ref{tab:fst}, the performance of \cite{boulch2022poco} drops by roughly fourfold going from testing on synthetic data to real scans.\\ 
\noindent \textbf{Limited input size generalization}\quad As illustrated in table \ref{tab:size}, when presented with testing point clouds that are larger (\eg 10k) than the training size (3k), the performance does not scale and even deteriorates compared to testing on the training size.\\ 
\noindent \textbf{Underfitting the input}\quad There is no explicit guaranty the input points are at the predicted shape function level set from a single forward pass. We hypothesise that enforcing this constraint under well chosen regularizations could potentially lead to improvements at test time and a reduction of the generalization issues mentioned above. 

We want to formulate an efficient strategy to improve on these aspects of generalizable shape networks at test time, \ie reconstructing shapes by dynamically adjusting the network to the input point-cloud during testing. We note that we follow such a strategy as training large models from scratch can be resource intensive, while adaptation can be less costly and has been shown to yield considerable improvements in \eg domain adaptation and transfer learning literature. One possible solution \cite{tang2021sa} (SAC) is to finetune the network weights at test time, using the knowledge that input point cloud samples are on the desired surface as supervision. However, we find this strategy to be unstable as it can overfit on these samples. As shown throughout the results (Section \ref{sec:res}), it even exacerbates the performance of its initial baseline network in many cases. This instability highlights the necessity for regularization in this adaptation process. We need to balance the need to maintain the expressive capacity of the learned functions while ensuring stability and robustness. To this end we devise three important design choices:\\
\noindent $\bullet$ We restrain the hypothesis space of the adapted shape functions to be a Reproducing Kernel Hilbert Space (RKHS). By the Representer Theorem, the minimizer of our regularized empirical risk minimization (ERM) problem (Equation \ref{equ:erm}) emerges naturally as the solution of a Kernel Ridge Regression (KRR) problem.\\
\noindent $\bullet$ By using a Gaussian kernel in this context, we benefit from the universality properties of the associated (RKHS) \ie a hypothesis space rich enough to approximate any continuous function arbitrarily well. Thus, we avoid the difficulties arising from optimizing a large number of neural network parameters, as in SAC, while maintaining the expressive capacity necessary for effective shape modeling.\\
\noindent $\bullet$ We propose a strategy to solve our regularized ERM problem in a RHKS space adapted to the shape we would like to recover. This is possible thanks to the unique correspondence between RKHS and kernels. Hence, instead of relying on handmade fixed kernels, we learn the KRR hyperparameters using a loss function that avoids overfitting on the data term (Equation \ref{equ:loss2}).

In this regard, approximate KRR solvers based on Nystr\"{o}m samples (Nystr\"{o}m KRR) \cite{meanti2020kernel} can be computed efficiently and allow scaling in point cloud size. As shown in our ablation studies, defining this kernel in euclidean space seems insufficient to constrain the fitting problem. Hence we define it in the feature space of the pretrained convolutional shape network. To define the fitting data, we use the input point cloud with their inherent \textit{label}: being samples from the surface by default. We also augment the fitting data with additional samples paired with their shape network predictions as pseudo-labels. This augmentation ensures the fitting problem is well defined. We note that through this strategy, knowledge is being transferred from the network to our kernel regression through both features and pseudo-labels. 

\begin{figure}[t!]
\centering
\includegraphics[width=0.95\linewidth]{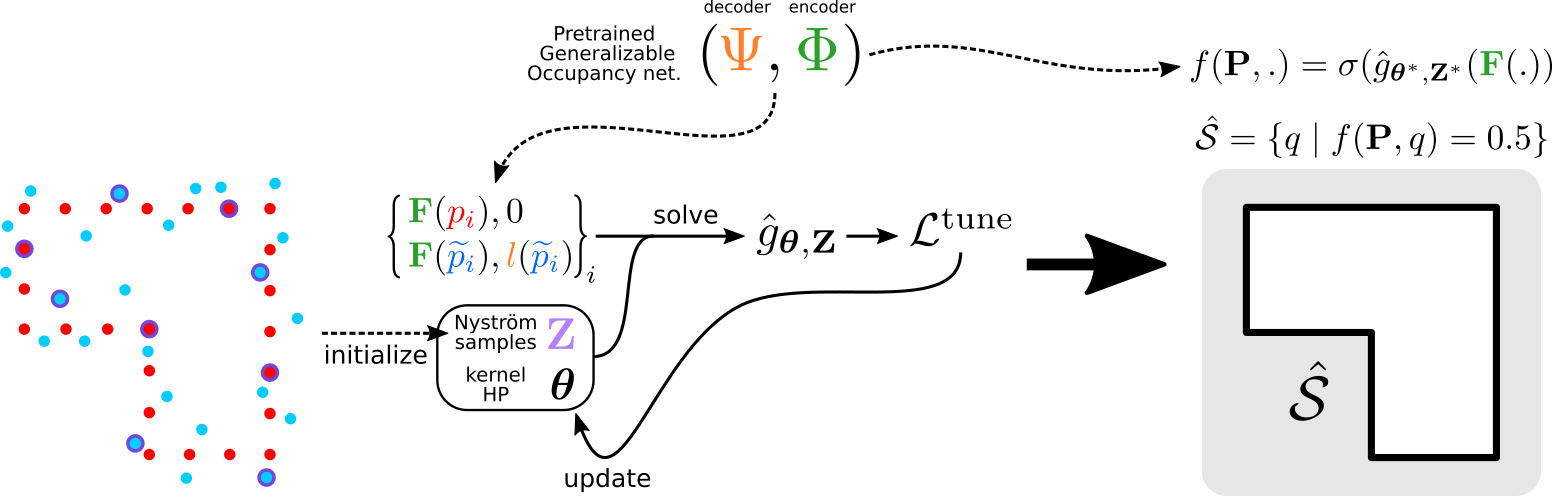}
\vspace{-10pt}
  \caption{Overview. Our method predicts an implicit shape function from a noisy unoriented input point cloud. We combine a cross-shape deep prior (Pretrained generalizable occupancy network) and an intra-shape adaptive Nystr\"{o}m Kernel Ridge Regression (NKRR) $g$ at test time. The latter learns to map \textcolor{Green}{network features} of the \textcolor{Red}{input points} and \textcolor{Cyan}{additional points} to the level-set and \textcolor{Orange}{network generated pseudo-labels}  respectively. The NKRR hyperparameters (\textcolor{Orchid}{$\textbf{Z}$},$\boldsymbol{\theta}$) are adjusted to the current shape.}  
  \label{fig:pipe}
\end{figure}

Until recently, exiting methods for KRR hyperparameter tuning, \eg cross-validation or grid search, placed strong limits on the flexibility and number of hyperparameters that can be tuned. The work of Meanti \etal \cite{meanti2022efficient} proposed lately a automated gradient based tuning of the hyperparameters of the KRR by minimizing a loss that prevents overfitting on the fitting data. This strategy allows the optimization of both the kernel hyperprameters and the Nystr\"{o}m samples, and proves beneficial in our context, \ie fitting a shape function  in feature space, as compared to baselines with fixed Nystr\"{o}m samples and/or kernel hyperparameters (See ablation in section \ref{sec:abl}). A visual summary of our method can be found in Figure \ref{fig:pipe}.  

To test our idea we devise experiments on real and synthetic data targeting generalization issues. Our approach improves consistently on the baseline networks, and outperforms other state-of-the-art methods and test time network finetuning based strategies. We also show that our approach can be applied successfully to more than one network. Our ablation studies showcase also the importance of all the various components of our method.  

%In Summary, our main contributions in this work are:\\
%\noindent $\bullet$ Achieving 3D reconstruction from point cloud by transfer learning from a preexisting deep neural network to a tunable Nystr\"{o}m KRR model operating in feature space, and optimizing the hyperparameters of the latter subsequently, Thus improving on the network's initial generalization ability. 
%\noindent $\bullet$  To the best of our knowledge, this is the first work that uses a tunable kernel method to reconstruct shapes from point clouds without requiring normals.  

\section{Related Work}

We review in this section previous work we deemed most relevant to our problem and contribution.
\vspace{-5pt}
\paragraph{Shape Representations in Deep Learning}
Shapes can be represented within deep learning either intrinsically or extrinsically. Intrinsic representations are a discrimination of the shape itself. When done explicitly, using \eg tetrahedral or polygonal meshes \cite{wang2018pixel2mesh,kato2018neural} or point clouds \cite{fan2017point}, the output topology in predefined thus bounding the variability of the shapes that can be generated. Among other forms of intrinsic representations, 2D patches \cite{groueix2018papier,williams2019deep,deprelle2019learning} can prompt discontinuities, whilst the simplicity of shape primitives such as cuboids \cite{abstractionTulsiani17,zou20173d}, planes \cite{liu2018planenet} and Gaussians \cite{genova2019learning} limits their expressiveness. Differently, extrinsic shape representations model the entire space containing the scene or object of interest. Voxel grids \cite{wu20153d,wu2016learning} are the most popular one being the direct extension of 2D pixels to 3D domain. However, their capacity is limited by their cubic resolution memory cost. Sparse representations such as octrees \cite{riegler2017octnet,tatarchenko2017octree,wang2017cnn} can alleviate this issue to some extent. 

\vspace{-5pt}
\paragraph{Implicit Neural Shape Representations}
Implicit neural representations (INRs) emerged recently as a major medium for modelling extrinsic shape and radiance fields (\eg \cite{mildenhall2020nerf,yariv2021volume,wang2021neus,jain2021dreamfields,chan2022efficient}). They overcome many of the limitations of the aforementioned classical representations thanks to their ability to represent shapes with arbitrary topologies at virtually infinite resolution. They are usually parameterised with MLPs mapping spatial locations or features to \eg occupancy \cite{mescheder2019occupancy}, signed \cite{park2019deepsdf} or unsigned \cite{chibane2020neural,Zhou2022CAP-UDF} distances relative to the target shape. %Shape INR models can be trained fully supervisedly using groundtruth labelled 3D samples \cite{park2019deepsdf,mescheder2019occupancy,chibane2020neural}, or through weaker signals such as 2D color and depth images \cite{niemeyer2020differentiable,kellnhofer2021neural,NEURIPS2020_1a77befc} or segmentaion masks \cite{lin2020sdf}, notably through differentiable rendering \cite{jiang2020sdfdiff,NEURIPS2019_bdf3fd65,jiang2020sdfdiff}. 
The level-set of the inferred field from these MLPs can be rendered through ray marching \cite{hart1996sphere}, or tessellated into an explicit shape using \eg Marching Cubes \cite{lorensen1987marching}. Another noteworthy branch of work builds hybrid implicit/explicit representations \cite{palmer2022deepcurrents,chen2020bsp,deng2020cvxnet,yavartanoo20213dias} based mostly on differentiable space partitioning. In order to represent collections of shape simultaneously, implicit neural models require conditioning mechanisms. These include feature and latent code concatenation, batch normalization, hypernetworks \cite{sitzmann2020implicit,NEURIPS2019_b5dc4e5d,sitzmann2021light,wang2021metaavatar,chen2022transinr} and gradient-based meta-learning \cite{ouasfi2022few,sitzmann2020metasdf}. Concatenation based conditioning was first implemented using single global latent codes \cite{mescheder2019occupancy,chen2019learning,park2019deepsdf}, and further improved with the use of local features \cite{ouasfi2024mix,li2022learning,genova2020local,tretschk2020patchnets,takikawa2021neural,peng2020convolutional,chibane2020implicit,jiang2020local,erler2020points2surf}.

\vspace{-5pt}
\paragraph{Reconstruction from Point Cloud}
Classical approaches include combinatorical ones where the shape is defined through an input point cloud based space partitioning, through \eg alpha shapes \cite{bernardini1999ball} Voronoi diagrams \cite{amenta2001power} or triangulation \cite{cazals2006delaunay,liu2020meshing,rakotosaona2021differentiable}. On the other hand, the input samples can be used to define an implicit function whose zero level set represents the target shape, using global smoothing priors \cite{williams2022neural,lin2022surface,williams2021neural} \eg radial basis function \cite{carr2001reconstruction} and Gaussian kernel fitting \cite{scholkopf2004kernel}, local smoothing priors such as moving least squares \cite{mercier2022moving,guennebaud2007algebraic,kolluri2008provably,liu2021deep}, or by solving a boundary conditioned Poisson equation \cite{kazhdan2013screened}. %We note that while the work in \cite{williams2022neural,williams2021neural} uses kernel regression similarly to ours, they require oriented point clouds as input, while our method does not require normal information. 
The recent literature proposes to parameterise these implicit functions with deep neural networks and learn their parameters with gradient descent, either in a supervised or unsupervised manner.

\vspace{-5pt}
\paragraph{Unsupervised Implicit Neural Reconstruction}
A neural network is typically fitted to the input point cloud without extra information. Regularizations can improve the convergence such as the spatial gradient constraint based on the Eikonal equation introduced by Gropp \etal \cite{gropp2020implicit}, a spatial divergence constraint as in \cite{ben2022digs}, Lipschitz regularization on the network \cite{liu2022learning}. Atzmon \etal learns an SDF from unsigned distances \cite{atzmon2020sal}, and further supervises the spatial gradient of the function with normals \cite{atzmon2020sald}. Ma \etal \cite{ma2020neural} expresses the nearest point on the surface as a function of the neural signed distance and its gradient. They also leverage 
self-supervised local priors to deal with very sparse inputs \cite{ma2022reconstructing} and improve generalization \cite{ma2022surface}. All of the aforementioned work benefits from efficient gradient computation through back-propagation in the neural network. Periodic activations were introduced in \cite{sitzmann2020implicit}. Lipman \cite{lipman2021phase} learns a function that converges to occupancy while its log transform
converges to a distance function. \cite{williams2021neural} learns infinitely wide shallow MLPs as random feature kernels using points and their normals.  
Most of the aforementioned methods present failures under sparse and noisy input due to the lack of supervision and data priors. Differently, we propose here to combine a data prior based method and self-supervised learning.   

\vspace{-5pt}
\paragraph{Supervised Implicit Neural Reconstruction}
Supervised methods assume a labeled training data corpus commonly in the form of dense samples with ground truth shape information. Auto-decoding methods \cite{li2022learning,park2019deepsdf,tretschk2020patchnets,jiang2020local,chabra2020deep} require test time optimization to be fitted to a new point cloud, which can take up to several seconds. Encoder-decoder based methods enable fast feed forward inference. Introduced first in this respect, Pooling-based set encoders  \cite{mescheder2019occupancy,chen2019learning,genova2020local} such as PointNet \cite{qi2017pointnet} have been shown to underfit the context. Convolutional encoders yield state-of-the-art performances. They use local features either defined in explicit volumes and planes \cite{peng2020convolutional,chibane2020implicit,lionar2021dynamic,peng2021shape} or solely at the input points \cite{boulch2022poco}. Ouasfi \etal \cite{ouasfi2024mix} proposed concurrently the first convolution-free fast feed forward generalizable model. Peng \etal \cite{peng2021shape} proposed a differentiable Poisson solving layer that converts predicted normals into an indicator function grid efficiently. However, it is limited to small scenes due to the cubic memory requirement in grid resolution. Most of these methods still suffer from generalization issues. Concurrently, the work in \cite{williams2022neural,huang2023neural} proposes to build generalizable models where the implicit decoder is a kernel regression. Differently, we advocate a strategy to improve the generalization of any preexisting pretrained reconstruction network.

\section{Method}
\label{sec:method}

%\vspace{-10pt}

Our input is a noisy unoriented point cloud $\mathbf{P}\subset{\mathbb R}^{3\times N_p}$. Our objective is to recover a 3D reconstruction, \ie the shape surface $\mathcal{S}$ that best explains this observation, where the input point cloud elements approximate noisy samples from $\mathcal{S}$. 

To achieve this, we model an implicit function $f$ that predicts occupancy values relative to a target shape $\mathcal{S}$ at any queried euclidean space location $q \in {\mathbb R}^{3}$, given the input point cloud $\mathbf{P}$, \ie $f(\mathbf{P},q)=1$ if $q$ is inside the shape, and $0$ otherwise. The inferred shape $\hat{\mathcal{S}}$ can then be obtained as a level set of the occupancy field inferred through $f$:
\begin{equation}
\hat{\mathcal{S}} = \{ q\in\mathbb{R}^3 \mid f(\mathbf{P},q) = 0.5\}.
\end{equation}
In practice, an explicit triangle mesh for $\hat{\mathcal{S}}$ is extracted using the Marching Cubes \cite{lorensen1987marching} algorithm.

\subsection{Feedforward Generalizable Model}

The staple back-bone models for feedforward generalizable implicit shape reconstruction from point cloud (\eg \cite{peng2020convolutional,chibane2020implicit,boulch2022poco}) have typically a two-stage architecture. First, a deep convolutional network (usually a U-Net) builds features in $\mathbb{R}^{N_F}$ from the input point cloud: 
\begin{equation}
    \mathbf{F} = \Phi(\mathbf{P}).
\label{eq:enc}
\end{equation}
Features $\mathbf{F}$ can be extrinsic or intrinsic. For instance in \cite{peng2020convolutional,chibane2020implicit}, the ConvNet builds an extrinsic 2D or 3D explicit feature grid representing space around the shape. More recently, Boulch \etal \cite{boulch2022poco} argued against this strategy,  contending that voxel centers may be far from the input point cloud locations, and are sampled uniformly while they should be more focused near the surface of interest. They hence define features intrinsically, \ie only at the input point cloud locations. In both strategies, the ConvNet offers transnational equivariance and generalization ability. It is worth noting that it is also naturally endowed with a locality mechanism, which is crucial for performance with fast inference times, as only a single forward pass in the encoder is needed to build the feature space. Conversely, other generalizable models lacking locality in their encoders, such as the work in \cite{erler2020points2surf}, enforce locality through local patch inputs. They hence require as many encoder forward passes as there are query points. This naturally results in significantly increased inference times compared the the aforementioned competition that we build on.  

Given a 3D query point $q$, an implicit decoder, typically an MLP, maps the feature to the occupancy value:
\begin{align}
\text{occ}(q) &= \sigma(\Psi(\mathbf{F}(q)))\\
&= \sigma(l(q)),
\label{eq:dec}
\end{align}
where $\sigma(.)$ represents the Sigmoid activation and $l(.)$ a logit function. 
In the case of extrinsic feature networks \cite{peng2020convolutional,chibane2020implicit,peng2021shape}, feature $\mathbf{F}(q)$ is obtained by linear interpolation (trilinear in the case of 3D feature grids, and bilinear for 2D grids). For intrinsic feature networks \cite{boulch2022poco}, the feature is pooled order-invariantly from the nearest points of the input point cloud.  Without loss of generality we build in this work on the model in \cite{boulch2022poco}, as it has been shown to achieve state-of-the-art performances on several benchmarks.

\subsection{Kernel Ridge Regression as a Shape Function in Feature Space}

Existing pretrained generalizable feedforward models for reconstruction from point cloud suffer from generalization limitations. As they are typically trained on synthetic shape datasets (\eg ShapeNet \cite{shapenet}) with full supervision, using a predefined point cloud size and simulated noise, their performance reduces when tested on point clouds that are different from the training ones regarding properties such as density, and whether they are real or simulated. Such shortcomings of an instance of these models are showcased in Tables \ref{tab:size} and \ref{tab:scan}. We design henceforth an efficient mechanism that can fill this performance gap, while benefiting from the available shape labels at test time: The fact that the input point cloud is made of samples that belong to the surface we wish to reconstruct.   

We consider the following training data samples:
\begin{equation}
    {\{(x_i,y_i)\}}_{i=1}^{n}={\{(\mathbf{F}(p_i),0)\}}_{i=1}^{N_p}\cup {\{(\mathbf{F}(\widetilde{p}_i),l(\widetilde{p}_i))\}}_{i=1}^{N_p},
\label{equ:samples}    
\end{equation}
where $(x_i,y_i) \in \mathbb{R}^{N_F}\times\mathbb{R}$ and  $n=2\times N_p$. $\{p_i\}$ are the samples of the input point cloud $\mathbf{P}$. $\{\widetilde{p}_i\}$ are additional samples around this point cloud, that can be automatically generated following \eg the strategy in \cite{ma2020neural}. Features $\mathbf{F}(.)$ and occupancy logits $l(.)$ are obtained from the pretrained convolutional occupancy model (See Equations \ref{eq:enc} and \ref{eq:dec}).  We note that these data samples are corrupted by a combination of observation noise in the input point cloud (points $p_i$ not being exactly at the surface) and prediction errors of the initial model (pseudo-labels $l(\widetilde{p}_i)$ being inaccurate). 

We approximate the mapping from deep shape features $x_i$ to occupancy logits $y_i$ using a non-linear non-parametric regression $g(.)$ to provide spatial regularization while favouring flexibility, \ie without making strong assumptions on the model. In particular, we use Kernel Ridge Regression (KRR), where the hypothesis space of the mapping is a reproducing kernel Hilbert space $\mathcal{H}$. Associated to this space is a kernel function 
$k_{\boldsymbol{\theta}}: \mathbb{R}^{N_F}\times\mathbb{R}^{N_F}\rightarrow \mathbb{R}$, that depends on hyperparameters $\boldsymbol{\theta}$. To ensure the minimisation of the square error of the mapping over the training samples is well defined, a regularization is typically needed, with a weight $\lambda$:
%\begin{equation}
%\hat{g}_\theta = \underset{g}{\arg\min}\sum_i^n\|g(x_i)-y_i\|^2 + %\lambda\|g\|^2_{\mathcal{H}}.
%\label{equ:min}
%\end{equation}
\begin{gather}
\hat{g}_{\boldsymbol{\theta}} = \underset{g\in\mathcal{H} }{\arg\min}\|g(\mathbf{X})-\mathbf{Y}\|^2 + \lambda\|g\|^2_{\mathcal{H}},\label{equ:min}\\
\text{ where } \quad 
\mathbf{X} = \{x_i\}_{i=1}^{n}, \quad \mathbf{Y} = \{y_i\}_{i=1}^{n}.
\label{equ:erm}
\end{gather}

The unique solution to this problem is expensive to compute. For efficiency we follow the approximation to KRR that considers a lower subspace of dimension $m<<n$ \cite{williams2000using}. This allows our method to scale to large input point clouds. The approximation uses $m$ inducing features $\textbf{Z}=\{z_j\}_{j=1}^{m}$, also known as Nystr\"{o}m Samples, where $z_j \in \mathbb{R}^{N_F}$. 

Finally, the unique solution to the minimization in Equation \ref{equ:min} writes \cite{meanti2020kernel}:
\begin{gather}
\hat{g}_{\boldsymbol{\theta}, \textbf{Z}} =\sum_{j=1}^{m} \beta_{j} k_{\boldsymbol{\theta}}\left(\cdot, z_{j}\right),\label{equ:krr}\\
\text { where }  \quad 
\boldsymbol{\beta} =\left(\textbf{K}_{n m}^{\top} \textbf{K}_{n m}+\lambda n \textbf{K}_{m m}\right)^{-1} \textbf{K}_{n m}^{\top} \textbf{Y}.
\label{equ:beta}
\end{gather}
Here, $\left(\textbf{K}_{n m}\right)_{i, j}=k_{\boldsymbol{\theta}}\left(x_{i}, z_{j}\right)$ and $\left(\textbf{K}_{m m}\right)_{i, j}=k_{\boldsymbol{\theta}}\left(z_{i}, z_{j}\right)$, where $i\in \llbracket 1,n \rrbracket$ and  $j\in \llbracket 1,m \rrbracket$. This solution can be computed efficiently through \eg the Falcon solver \cite{meanti2020kernel}.   

\subsection{Learning the Shape Function Hyperparameters}

Next, based on \cite{meanti2022efficient}, we propose to automatically learn the hyperparameters of our Nystr\"{o}m KRR, namely the kernel hyperparameters $\boldsymbol{\theta}$ and the Nystr\"{o}m samples $\mathbf{Z}$,  using gradient descent. This shape specific tuning allows for a larger hyperparameter space and using more expressive kernels. The optimization consists in minimizing an upper bound $\mathcal{L}^{\text {tune }}$ of the ideal objective, \ie the expectation over all possible data points of the squared error of the model. This allows to avoid overfitting on the training set:
\begin{gather}
\mathcal{L}^{\text {tune }}= \frac{2}{n} \operatorname{Tr}\left((\widetilde{\mathbf{K}}+n \lambda \mathbf{K})^{-1} \widetilde{\mathbf{K}}\right) +\frac{2}{n \lambda} \operatorname{Tr}(\mathbf{K}-\widetilde{\mathbf{K}})\mathcal{L}^{\text {data}}(\hat{g}_{\boldsymbol{\theta}, \mathbf{Z}}) + 2\mathcal{L}^{\text {data}}(\hat{g}_{\boldsymbol{\theta}, \mathbf{Z}}),\label{equ:loss1}\\
\text{where} \quad 
\mathcal{L}^{\text {data}}(\hat{g}_{\boldsymbol{\theta}, \mathbf{Z}})= \frac{1}{n}\|\hat{g}_{\boldsymbol{\theta}, \mathbf{Z}}(\mathbf{X})-\mathbf{Y}\|^2+\lambda\|\hat{g}_{\boldsymbol{\theta}, \mathbf{Z}}\|_{\mathcal{H}}^2.
\label{equ:loss2}
\end{gather}

$\widetilde{\mathbf{K}}= \mathbf{K}_{nm}\mathbf{K}^{\dagger}_{mm}\mathbf{K}^{\top}_{nm}$
and $^{\dagger}$ symbolizes the Moore-Penrose matrix inverse. 
The upper bound is made, in this order, of he Nystr\"{o}m  penalty attempting to make the Nystr\"{o}m kernel
$\widetilde{\mathbf{K}}$ closer to the complete kernel $\mathbf{K}$, a second term penalizing overly complex kernels, and finally a data fit term. We note that the authors in \cite{meanti2022efficient} derived this loss by replacing expectations with their empirical counterparts. They also propose to use the Hutchinson approximation to compute matrix traces for efficiency.

Aftre $N$ gradient descent steps (as illustrated in Algorithm \ref{alg:train}), we select hyperparameters $\boldsymbol{\theta}^*$ and $\mathbf{Z}^*$ that minimize the squared error of the model over the training data following \cite{meanti2022efficient}. For a given query point q, our final implicit occupancy function can be expressed as the sigmoid activated optimal Nystr\"{o}m KRR in feature space:
\begin{equation}
    f(\mathbf{P},q) = \sigma(\hat{g}_{\boldsymbol{\theta}^*,\mathbf{Z}^*}(\mathbf{F}(q)))
\end{equation}

\begin{algorithm}
\small
\begin{algorithmic}
\Require Point cloud $\textbf{P}$, pretrained occupancy network $(\Psi,\Phi)$, learning rate $\alpha$ 
\Ensure optimal kernel hyperparameters $\boldsymbol{\theta}^*$ and Nystr\"{o}m samples $\textbf{Z}^*$   
\State $\widetilde{\textbf{P}} = \text{upsample}(\textbf{P})$
\State $\textbf{X} = \Phi([\textbf{P},\widetilde{\textbf{P}}])$
\State $\textbf{Y} = [\textbf{0}, \Psi \circ \Phi(\widetilde{\textbf{P}})]$
\State initialize $\textbf{Z}$ as $m$ random features from \textbf{X}, initialize $\boldsymbol{\theta}$
\For {$N$ times}
\State compute $\hat{g}_{\boldsymbol{\theta},\mathbf{Z}}$ from $\textbf{X}$, $\textbf{Y}$ (Equations \ref{equ:krr} and \ref{equ:beta})
\State compute $\mathcal{L}^{\text{tune}}$ for $\hat{g}_{\boldsymbol{\theta},\mathbf{Z}}$, $\textbf{X}$ and $\textbf{Y}$ (Equations \ref{equ:loss1} and \ref{equ:loss2})
\State $(\boldsymbol{\theta},\textbf{Z}) \leftarrow (\boldsymbol{\theta},\textbf{Z}) - \alpha \nabla_{\boldsymbol{\theta},\mathbf{Z}} \mathcal{L}^{\text{tune}}$
\EndFor
\end{algorithmic}
\caption{\small The training procedure of our method.}
\label{alg:train}
\end{algorithm}

\section{Results}
\label{sec:res}
In this section, we evaluate our method using real and synthetic data both quantitatively and qualitatively. As our method builds on Poco \cite{boulch2022poco}, we compare our results to this baseline. We also compare to other state-of-the-art convolutional occupancy networks, such as Conv \cite{peng2020convolutional} and SAP \cite{peng2021shape}. We compare to the strategy based on test-time finetuning of a convolutional occupancy network (Conv \cite{peng2020convolutional} ) in SAC \cite{tang2021sa}. Additionally, we compare to the strategy based on fitting an MLP to the point cloud in NP \cite{ma2020neural}. We compare also to classical learning-free Poisson reconstruction (SPSR \cite{kazhdan2013screened}) where input point normals are estimated with PCA. 
For state-of-the-art convolutional occupancy networks Poco, Conv and SAP, we use their official publicly available pretrained models as trained on the same training split of ShapeNet \cite{shapenet} with 3k sized point clouds and noise of variance 0.005. We use the publicly available official implementations of SAP and NP. We present also an ablative analysis of the components of our method in Section \ref{sec:abl}. We provide additional results in the supplementary material. We note that from a computational stand point, for a 10k sized input point cloud, our method takes roughly 10 seconds to converge on a NVIDIA RTX A6000 GPU. The other test-time tuning based alternatives we compare to here, namely SAC and NP take roughly 1 and 5 minutes respectively.    

\subsection{Implementation Details}
 We run all the experiments for 100 epochs using Adam optimizer with learning rate 0.1.
We use the Falcon Library to implement the tuning of Nystr\"{o}m KRR in feature space. Regarding the input size and feature dimensions, We experimented with $N_P = 10k$ (also 500 and 3k in ablation) and $N_F = 32$ and $m = 500$ inducing points.
% We used the Gaussian kernel with a single length-scale for each data-dimension that we initially set at 0.001.

\subsection{Metrics}

Following seminal work, we evaluate our method and the competition \wrt the ground truth using standard metrics for the 3D reconstruction task. Namely, the volumetric \textbf{Intersection over Union (IoU)} when ground truth meshes are available, the L1 \textbf{Chamfer Distance
CD$_1$} ($\times10^{2}$) and euclidean distance based \textbf{F-Score (FS)} when ground truth points are available, and finally \textbf{Normal Consistency (NC)} when ground truth normals are available. We detail the expressions of these metrics in the supplementary material.

\subsection{Datasets}

\textbf{ShapeNet} \cite{shapenet} consists of various instances of 13 different synthetic 3D object classes. Following \cite{williams2021neural}, we use the first 20 shapes from the test split of each class for evaluation. We generate inputs of different sizes (10k, 3k, 500) while adding gaussian noise of standard deviation 0.005 following the literature (\eg \cite{boulch2022poco,peng2020convolutional}). % We use this dataset mainly to asses  generalization across various point cloud sizes and to conduct an ablation study for our method.  
\textbf{ScanNet v2}\cite{dai2017scannet} is a challenging real dataset containing 1513 room scans as captured with an RBG-D camera. We use 100 scans from the test split for evaluation. We sample 10k points from the scans as input. %We use this dataset as an extreme generalization test of models trained on synthetic objects. 
\textbf{Faust} \cite{Bogo:CVPR:2014} consists of real scans of 10 human body identities in 10 different poses. We use the entire 100 scans for evaluation. We sample 10k points from the scans as inputs. We use the provided mesh registrations to compute IoU. %We use this dataset as an additional generalization test to another type of real data for models trained on synthetic objects. 

\begin{figure}[t!]
\centering
\includegraphics[width=.9\linewidth]{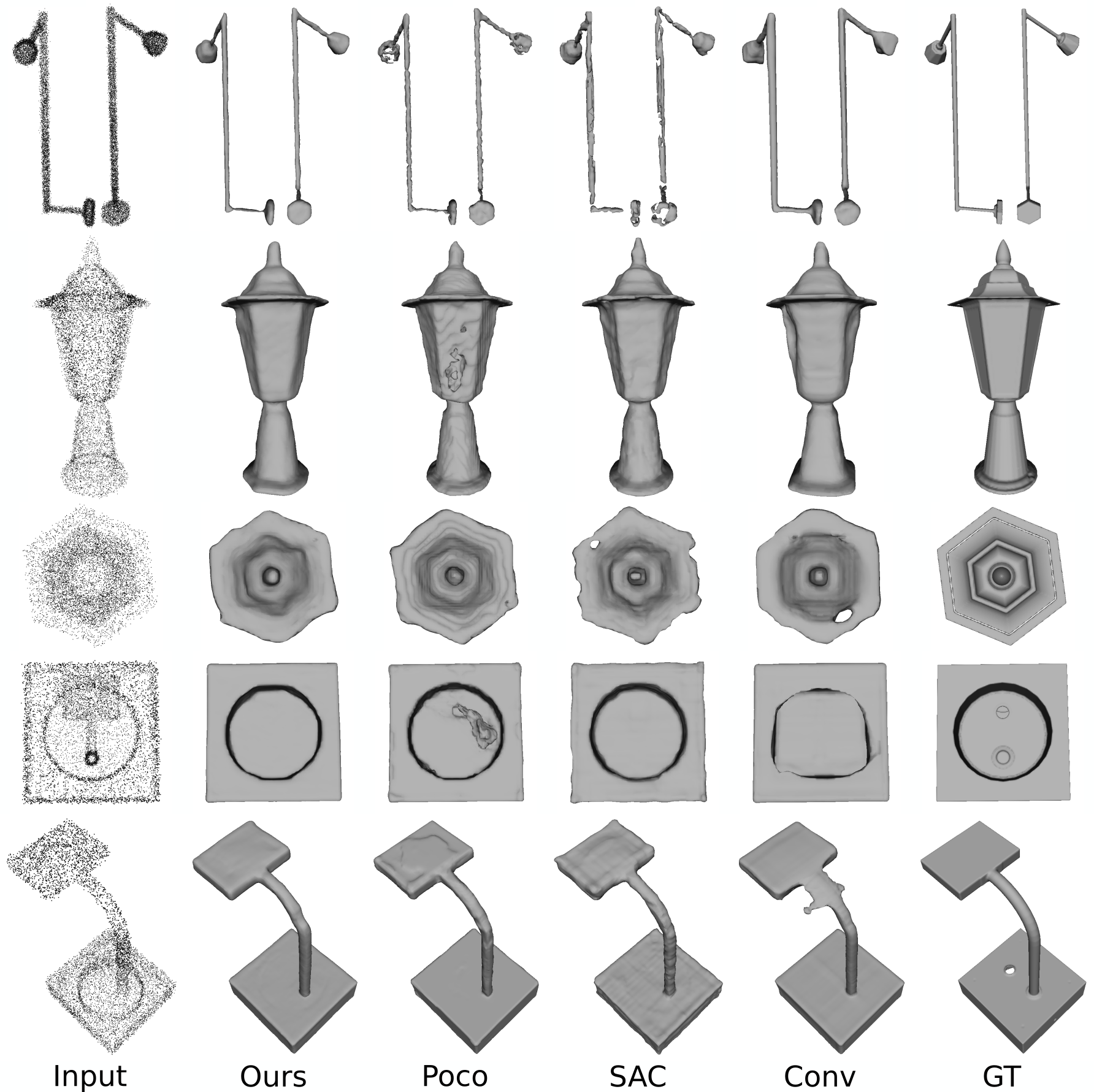}
\vspace{-10pt}
  \caption{Qualitative comparison of ShapeNet reconstructions.}
  \label{fig:sn}
\end{figure}

\subsection{Object Level Reconstruction}

\begin{table}[h!]
\begin{minipage}{.45\linewidth}
\centering
\caption{IoU $\uparrow$ of ShapeNet reconstruction.}
    \begin{tabular}{@{}lccccc@{}}
        \toprule
        {} & Conv & SAC & NP & Poco & Ours \\
%        Classes & & & & & \\
        \midrule
        Chair & 0,87 & 0,88 & 0,67 & 0,90 & 0,91 \\
        Lamp & 0,77 & 0,75 & 0,64 & 0,80 & 0,87 \\
        Table & 0,89 & 0,79 & 0,71 & 0,89 & 0,91 \\
        \midrule
        Mean & 0,84 & 0,81 & 0,67 & 0,86 & \textbf{0,90} \\
        \bottomrule
        \label{tab:sniou}
    \end{tabular}
\end{minipage}
 \hspace{30pt}
\begin{minipage}{.45\linewidth}
\centering
\caption{CD$_1$ $\downarrow$ of ShapeNet reconstruction.}
    \begin{tabular}{@{}lccccc@{}}
        \toprule
        {} & Conv & SAC & NP & Poco & Ours \\
%        Classes & & & & & \\
        \midrule
        Chair & 0,41 & 0,39 & 0,56 & 0,42 & 0,34 \\
        Lamp & 0,47 & 0,44 & 0,72 & 0,56 & 0,42 \\
        Table & 0,35 & 0,56 & 0,62 & 0,38 & 0,30 \\
        \midrule
        Mean & 0,41 & 0,46 & 0,63 & 0,45 & \textbf{0,35} \\
        \bottomrule
    \label{tab:sncd}
    \end{tabular}
\end{minipage}
\end{table}
\vspace{-15pt} 
\begin{table}[h!]
\begin{minipage}{.45\linewidth}
\centering
\caption{FS $\uparrow$ of ShapeNet reconstruction.}
    \begin{tabular}{@{}lccccc@{}}
        \toprule
        {} & Conv & SAC & NP & Poco & Ours \\
%        Classes & & & & & \\
        \midrule
        Chair & 0,95 & 0,98 & 0,87 & 0,98 & 0,99 \\
        Lamp & 0,92 & 0,94 & 0,83 & 0,95 & 0,96 \\
        Table & 0,98 & 0,92 & 0,80 & 0,98 & 0,99 \\
        \midrule
        Mean & 0,95 & 0,95 & 0,83 & 0,97 & \textbf{0,98} \\
        \bottomrule
        \label{tab:snfs}
    \end{tabular}
\end{minipage}
 \hspace{30pt}
\begin{minipage}{.45\linewidth}
\centering
\caption{NC $\uparrow$ of ShapeNet reconstruction.}
    \begin{tabular}{@{}lccccc@{}}
        \toprule
        {} & Conv & SAC & NP & Poco & Ours \\
%        Classes & & & & & \\
        \midrule
        Chair & 0,95 & 0,95 & 0,88 & 0,95 & 0,96 \\
        Lamp & 0,9 & 0,89 & 0,90 & 0,90 & 0,92 \\
        Table & 0,96 & 0,93 & 0,78 & 0,96 & 0,97 \\
        \midrule
        Mean & 0,94 & 0,92 & 0,85 & 0,94 & \textbf{0,95} \\
        \bottomrule
        \label{tab:snnc}
    \end{tabular}
\end{minipage}
\end{table}
\vspace{-10pt} 

We show here reconstruction results from 10k sized noisy point clouds of classes Chair, Table and Lamp of ShapeNet. Convolutional occupancy models were trained with 3k sized noisy inputs. We report a comparison on IoU in Table \ref{tab:sniou}, CD$_1$ in Table \ref{tab:sncd}, FS in Table \ref{tab:snfs}, and NC in Table \ref{tab:snnc}. 
Figures \ref{fig:sn} and \ref{fig:spsr} shows qualitative comparisons. 
Our method improves consistently across all metrics over its baseline Poco, as well as the other strategies. Methods that benefit from deep data priors (Conv, SAC, Poco, Ours) outperform data prior free \textit{overfitting} (NP). We note also that these results show that the finetuning strategy SAC is not stable and does not improve consistently on its deep convolutional occupancy network Conv. This is also confirmed in the visual results in Figure\ref{fig:sn}, where the finetuning can lead to breakage in the reconstruction. Our method delivers more robust reconstruction generally with less holes and artifacts. As it can be seen in the same figure, it can particularly recover from failure cases of its baseline Poco. We believe that many of the failures of the baseline Poco are due to its lack of generalization \wrt the input size. Figure \ref{fig:spsr} shows additional visual comparisons to SPSR. We provide a numerical comparison to this classical method in the supplementary material. 

\subsection{Generalization to Real Scene Scans}

\begin{figure}[h!]
\centering
\includegraphics[width=0.9\linewidth]{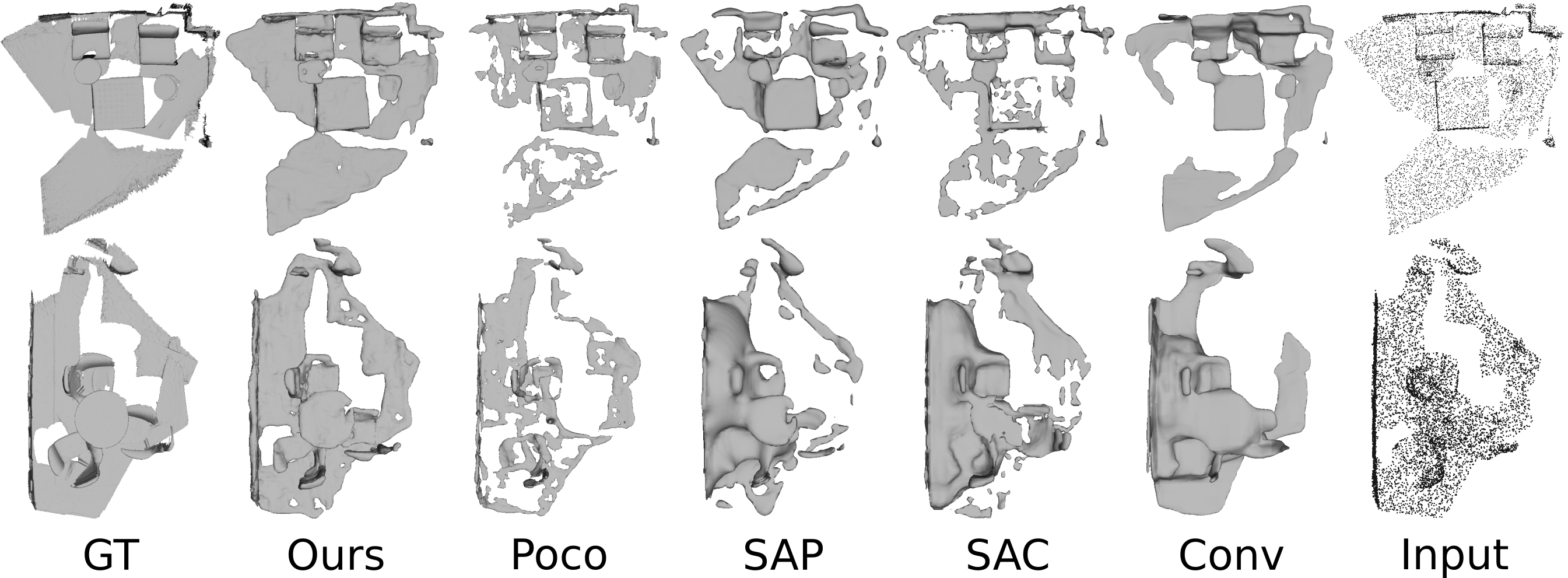}
\vspace{-10pt}
  \caption{Qualitative comparison of ScanNet v2 reconstructions.}
  \label{fig:scan}
\end{figure}

\begin{table}[h!]
\begin{minipage}{.45\linewidth}
\centering
\caption{ScanNet reconstruction.}
    \begin{tabular}{@{}lccc@{}}
        \toprule
        {} & CD$_1${$\downarrow$} & NC$\uparrow$ & FS$\uparrow$ \\
%        Methods & & & \\
        \midrule
        SPSR & 2.27 & 0.74 & 0.68 \\
        SAP & 1,12 & 0,73 & 0,67 \\
        Conv & 1,22 & 0,71 & 0,60 \\
        SAC & 0,97 & 0,76 & 0,77 \\
        Poco & 1,11 & 0,77 & 0,79 \\
        Ours (Poco) & 0,94 & \textbf{0,79} & \textbf{0,80} \\
        Ours (Conv) & \textbf{0,70} & 0,77 & 0,79 \\
        \bottomrule
        \label{tab:scan}
    \end{tabular}
\end{minipage}
 \hspace{20pt}
\begin{minipage}{.45\linewidth}
\centering
\caption{Faust reconstruction.}
    \begin{tabular}{@{}lcccc@{}}
        \toprule
        {}      & IoU$\uparrow$  & CD$_1${$\downarrow$} & NC$\uparrow$ & FS$\uparrow$ \\
%        Methods & & & & \\
        \midrule
        SPSR        & - & 0.48 & 0.91 & 0.91\\
        SAP         & 0,90 & 0,29 & 0,93 & 0,98 \\
        Conv        & 0,85 & 0,41 & 0,92 & 0,95 \\
        SAC     & 0,77  & 0,30 & 0,92	& 0,95   \\
        Poco        & 0,90  & 0,32 & 0,93 & 0,97 \\
        Ours        & \textbf{0,92}  & \textbf{0,26} & \textbf{0,95} & \textbf{0,99} \\
        \bottomrule
        \label{tab:fst}
    \end{tabular}
\end{minipage}
\end{table}

%\begin{table}[h!]
%    \centering
%    \caption{ScanNet reconstruction}
%    \begin{tabular}{@{}lccc@{}}
%        \toprule
%        {} & CD$_1${$\downarrow$} & NC$\uparrow$ & FS$\uparrow$ \\
%        Methods & & & \\
%        \midrule
%        SAP & 1,12 & 0,73 & 0,67 \\
%        Conv & 1,22 & 0,71 & 0,60 \\
%        SAC & 0,97 & 0,76 & 0,77 \\
%        Poco & 1,11 & 0,77 & 0,79 \\
%        Ours & \textbf{0,94} & \textbf{0,79} & \textbf{0,80} \\
%        \bottomrule
%        \label{tab:scan}
%    \end{tabular}
%\end{table}

We evaluate here the generalization from synthetic object training to real scenes. We show reconstruction results from 10k points sampled from real scans of ScanNet v2 in Table \ref{tab:scan}. As meshes are not available we can not compute IoU here. Our method (Ours(Poco)) based on Poco outperforms the baseline and the competition, demonstrating superior generalization ability. Figure \ref{fig:scan} shows a few qualitative results. While we offer improved performances compared to the competition, we note that the quality of our reconstruction remains limited still is such a challenging setting. We also show that our method can be successfully applied to other existing occupancy networks, as combining our strategy with baseline Conv (Ours(Conv)) leads to a considerable improvement in its performance.

\subsection{Generalization to Real Articulated Shape Scans}

\begin{figure}[t!]
\vspace{-10pt}
\centering
\includegraphics[width=0.9\linewidth]{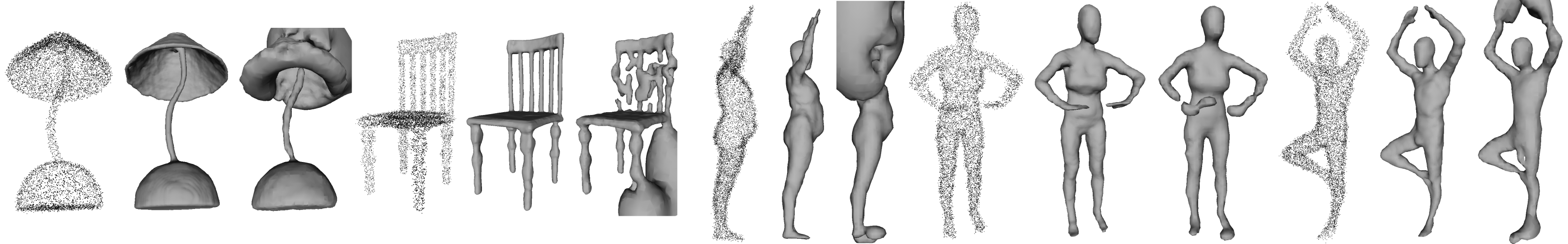}
\vspace{-10pt}
  \caption{Qualitative comparison to Screened Poisson Surface Reconstruction (SPSR). \textbf{Input / Ours / SPSR}, in this order.}
\label{fig:spsr}
\vspace{-10pt}
\end{figure}

\begin{wrapfigure}{l}{0.5\textwidth}
\centering
\includegraphics[width=\linewidth]{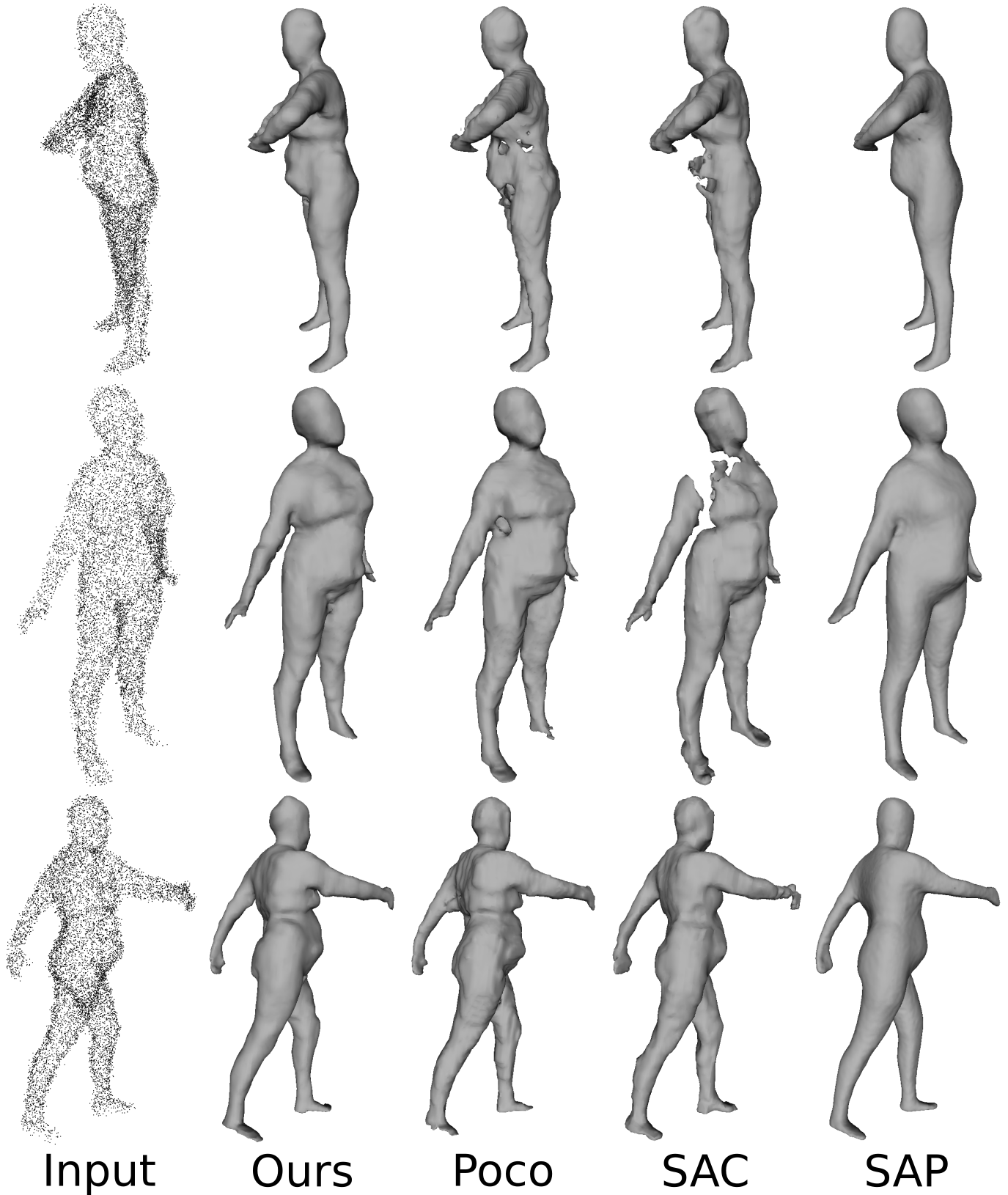}
\vspace{-15pt}
\caption{Qualitative comparison of Faust reconstructions.}
\label{fig:fst}
\vspace{-30pt}
\end{wrapfigure}

%\begin{figure}[h!]
%\centering
%\includegraphics[width=.6\linewidth]{fig/Faust.png}
%  \caption{Qualitative comparison of Faust reconstructions}
%  \label{fig:fst}
%\end{figure}

We evaluate here generalization from synthetic object training to real non-rigid human shapes. We show reconstruction results from 10k points sampled from real scans in Faust. As summarized in Table \ref{tab:fst}, both convolutional networks SAP and Poco offer good initial performances. Notice how the finetuning based SAC can improve of its baseline (Conv) CD$_1$ score while conversely worsening its IoU score. We outperform both our baseline and competing methods. Figure \ref{fig:fst} shows a visual comparison of the results, where we recover from some failures of our baseline (Poco), and show limitations of the finetuning strategy in SAC. Figure \ref{fig:spsr} shows additional visual comparisons to SPSR.  

%Qualitative results can be seen the supplementary material.

%\begin{table}[h!]
%    \centering
%    \caption{Faust reconstruction}
%    \begin{tabular}{@{}lcccc@{}}
%        \toprule
%        {}      & IoU$\uparrow$  & CD$_1${$\downarrow$} & NC$\uparrow$ & FS$\uparrow$ \\
%        Methods & & & & \\
%        \midrule
%        SAP         & 0,90 & 0,29 & 0,93 & 0,98 \\
%        Conv        & 0,85 & 0,41 & 0,92 & 0,95 \\
%        SAC     & 0,77  & 0,30 & 0,92	& 0,95   \\
%        Poco        & 0,90  & 0,32 & 0,93 & 0,97 \\
%        Ours        & \textbf{0,92}  & \textbf{0,26} & \textbf{0,95} & %\textbf{0,99} \\
%        \bottomrule
%        \label{tab:fst}
%    \end{tabular}
%\end{table}

\subsection{Ablative Analysis}
\label{sec:abl}

We conduct further ablative studies of our proposed method on the Lamp class of ShapeNet, as we found its shapes to be challenging and displaying many fine and complex features.  Table \ref{tab:size} shows the performance of both our method and our baseline network Poco \cite{boulch2022poco} for inputs of size 500, 3k and 10k. We note that Poco has been trained on 3k sized inputs. Notice first how the performance of Poco improves in CD$_1$ while deteriorating in IoU. This shows the performance of this model does not necessarily scale with input size. Our method on the other hand shows healthy consistent improvements over the baseline Poco, which also scale with input size. We notice also that the performance gap \wrt Poco is slightly tighter in the sparsest setup (500 points). We believe this is due to the inductive bias of kernel methods favouring smoothness, as such a prior can prove less effective for extremely sparse data. 

Table \ref{tab:mob} shows an ablation of the design choices of our method. Our performance drops as expected whether we disable the kernel hyperparameters $\boldsymbol{\theta}$ finetuning (w/o $\boldsymbol{\theta}$), the Nystr\"{o}m samples $\mathbf{Z}$ finetuning (w/o $\mathbf{Z}$). We note also the importance of learning the regression in feature space. We notice that reconstruction often fails when we attempt to solve the regression in euclidean space. Hence our w/o Feat. version uses a Fourier positional encoding to stabilize this baseline. However, this version still under performs compared to our final model. Disabling the tuning all together (w/o tune) results in a performance drop as well. Increasing the number of number of Nystr\"{o}m samples in this context improves the performance. 

Table \ref{tab:mob} provides also an ablation the number of Nystr\"{o}m samples $m$ while tuning (Ours). We find that generally the higher the number of inducing features $\textbf{Z}$ the better the performance, as it was the case without hyper-parameter optimization (w/o tune). This is expected as the approximation error of the Nystr\"{o}m KRR decreases with more inducing points. Additionally, more learnable inducing points implies more representation power for the KRR function. Increasing $m$ comes however with increased computation overhead due to both solving the Nystr\"{o}m KRR with a higher number of inducing features $\textbf{Z}$ and also learning these features. Increasing $m$ beyond a certain limit can also lead to overfitting. As our aim in this work is an efficient approach to improve generalization, we find a relatively small $m$ value (500) already sufficient to produce satisfactory shape predictions and offers a good performance/compute overhead trade-off.

\begin{table}[h!]
%\centering
\scalebox{0.97}{
\begin{minipage}[t]{0.45\linewidth}
\centering
    \caption{Ablation of input size.}
    \begin{tabular}{@{}lcccc@{}}
        \toprule
        {}      & IoU$\uparrow$  & CD$_1${$\downarrow$} & NC$\uparrow$ & FS$\uparrow$ \\
        %Methods & & & & \\
        \midrule
        Poco(500)         & 0,71	& 0,85	& \textbf{0,86}	& 0,84 \\
        Ours (500)     &  \textbf{0,73}	 & \textbf{0,80}	& 0,85	& \textbf{0,85} \\
        \midrule
        Poco(3k)        & 0,85	& 0,60	& 0,91	& \textbf{0,96} \\
        Ours (3k)       &  \textbf{0,86} & \textbf{0,52}	& \textbf{0,92} & 	\textbf{0,96} \\
        \midrule
        Poco(10k)     & 0,80	 & 0,56	& 0,90	& 0,95   \\
        Ours (10k)      &  \textbf{0,87}	& \textbf{0,46}	& \textbf{0,92} & 	\textbf{0,96} \\
        \bottomrule
        \label{tab:size}
    \end{tabular}
\end{minipage}
\hspace{10pt}
% \hfill
\begin{minipage}[t]{0.45\linewidth}
\centering
    \caption{Ablation of our model.}
    \begin{tabular}{@{}lcccc@{}}
        \toprule
        {}      & IoU$\uparrow$  & CD$_1${$\downarrow$} & NC$\uparrow$ & FS$\uparrow$ \\
        %Ours & & & & \\
        \midrule
        w/o tune ($m=1k$)         &0.84  &0.61   &\textbf{0.92}  &0.96 \\
        w/o tune ($m=5k$)      & \textbf{0.87}  &0.55  &\textbf{0.92}  &\textbf{0.97}\\
        w/o tune ($m=10k$)   &\textbf{0.87}  &\textbf{0.53}   &\textbf{0.92} &0.96\\
        \midrule
        w/o  Feat.       & 0.45   &2.23   &0.79  &0.58 \\
        w/o $\boldsymbol{\theta}$    &  0.86  & 0.52  &    0.92 & 0.95\\
       % \midrule
        w/o $\mathbf{Z}$       & 0.86 &   0.53  & 0.92 &   0.96 \\

        \midrule
        %Ours (10k)      &  \textbf{0,87}	& \textbf{0,46}	& \textbf{0,92} & 	\textbf{0,96} \\
        Ours ($m=500$)         &0.87  &0.42   &\textbf{0.92}  &0.96 \\
        Ours ($m=1000$)      & \textbf{0.88}  &\textbf{0.41}  &\textbf{0.92}  &\textbf{0.97}\\
        Ours ($m=5000$)   &0.86  &0.43   &0.91 &0.96\\
        \bottomrule
        \label{tab:mob}
    \end{tabular}
\end{minipage}
}
\vspace{-20pt}
\end{table}

Finally, we note that we chose to use an equal number of input and augmented samples (Equation \ref{equ:samples}) empirically as it gave the best overall results. It is possible that for high levels of noise, a higher number of augmented samples could prove beneficial. However, higher levels of noise for the input also imply less accurate predictions from the occupancy network, meaning less reliable pseudo-labels for the augmented samples. 

\vspace{-5pt}
\section{Limitations}
Extreme generalization from relatively sparse point cloud is still challenging for our method as well as the literature and we will seek further improvement in future work. Although it offers robustness, the inherent inductive bias in our method also favours smoothness, which can prove not ideal for representing the highest level of detail. Ablation of number of the NKRR Nystr\"{o}m samples $m$ in Table \ref{tab:mob} shows that the KRR can possibly suffer from overfitting for large values of $m$, hence setting this hyperparameter can be challenging. We provide visualizations and a short discussion of the behavior of our method under some baseline failures in the supplementary material.

\vspace{-5pt}
\section{Conclusion}
We presented a method for shape reconstruction from unoriented point cloud that combines the data prior of a preexisting deep reconstruction network and an efficient non-parametric interpolant. The resulting combination offers improved generalization over the network baseline, network test-time finetuning, and several state-of-the-art methods. We hope we can inspire more work in the direction of fine-tuning and transfer learning from preexisting feedforward occupancy networks, especially since there are many competing strategies that could be explored. For instance, while we propose here to improve the boundary decision by learning the shape function, an alternative strategy could be to tune solely the features instead. 

%\section*{References}
%\medskip

{\small
\bibliographystyle{ieee_fullname}
\bibliography{egbib}

\begin{thebibliography}{10}\itemsep=-1pt

\bibitem{amenta2001power}
Nina Amenta, Sunghee Choi, and Ravi~Krishna Kolluri.
\newblock The power crust, unions of balls, and the medial axis transform.
\newblock {\em CG}, 2001.

\bibitem{atzmon2020sal}
Matan Atzmon and Yaron Lipman.
\newblock Sal: Sign agnostic learning of shapes from raw data.
\newblock In {\em CVPR}, 2020.

\bibitem{atzmon2020sald}
Matan Atzmon and Yaron Lipman.
\newblock Sald: Sign agnostic learning with derivatives.
\newblock In {\em ICML}, 2020.

\bibitem{ben2022digs}
Yizhak Ben-Shabat, Chamin~Hewa Koneputugodage, and Stephen Gould.
\newblock Digs: Divergence guided shape implicit neural representation for
  unoriented point clouds.
\newblock In {\em Proceedings of the IEEE/CVF Conference on Computer Vision and
  Pattern Recognition}, pages 19323--19332, 2022.

\bibitem{bernardini1999ball}
Fausto Bernardini, Joshua Mittleman, Holly Rushmeier, Claudio Silva, and
  Gabriel Taubin.
\newblock The ball-pivoting algorithm for surface reconstruction.
\newblock {\em TVCG}, 1999.

\bibitem{Bogo:CVPR:2014}
Federica Bogo, Javier Romero, Matthew Loper, and Michael~J. Black.
\newblock {FAUST}: Dataset and evaluation for {3D} mesh registration.
\newblock In {\em CVPR}, 2014.

\bibitem{boulch2022poco}
Alexandre Boulch and Renaud Marlet.
\newblock Poco: Point convolution for surface reconstruction.
\newblock In {\em Proceedings of the IEEE/CVF Conference on Computer Vision and
  Pattern Recognition}, pages 6302--6314, 2022.

\bibitem{carr2001reconstruction}
Jonathan~C Carr, Richard~K Beatson, Jon~B Cherrie, Tim~J Mitchell, W~Richard
  Fright, Bruce~C McCallum, and Tim~R Evans.
\newblock Reconstruction and representation of 3d objects with radial basis
  functions.
\newblock In {\em SIGGRAPH}, 2001.

\bibitem{cazals2006delaunay}
Fr{\'e}d{\'e}ric Cazals and Joachim Giesen.
\newblock {\em Effective Computational Geometry for Curves and Surfaces}.
\newblock 2006.

\bibitem{chabra2020deep}
Rohan Chabra, Jan~E Lenssen, Eddy Ilg, Tanner Schmidt, Julian Straub, Steven
  Lovegrove, and Richard Newcombe.
\newblock Deep local shapes: Learning local sdf priors for detailed 3d
  reconstruction.
\newblock In {\em ECCV}, 2020.

\bibitem{chan2022efficient}
Eric~R Chan, Connor~Z Lin, Matthew~A Chan, Koki Nagano, Boxiao Pan, Shalini
  De~Mello, Orazio Gallo, Leonidas~J Guibas, Jonathan Tremblay, Sameh Khamis,
  et~al.
\newblock Efficient geometry-aware 3d generative adversarial networks.
\newblock In {\em Proceedings of the IEEE/CVF Conference on Computer Vision and
  Pattern Recognition}, pages 16123--16133, 2022.

\bibitem{shapenet}
Angel~X Chang, Thomas Funkhouser, Leonidas Guibas, Pat Hanrahan, Qixing Huang,
  Zimo Li, Silvio Savarese, Manolis Savva, Shuran Song, Hao Su, et~al.
\newblock Shapenet: An information-rich 3d model repository.
\newblock {\em arXiv preprint arXiv:1512.03012}, 2015.

\bibitem{chen2022transinr}
Yinbo Chen and Xiaolong Wang.
\newblock Transformers as meta-learners for implicit neural representations.
\newblock In {\em European Conference on Computer Vision}, 2022.

\bibitem{chen2020bsp}
Zhiqin Chen, Andrea Tagliasacchi, and Hao Zhang.
\newblock Bsp-net: Generating compact meshes via binary space partitioning.
\newblock In {\em Proceedings of the IEEE/CVF Conference on Computer Vision and
  Pattern Recognition}, 2020.

\bibitem{chen2019learning}
Zhiqin Chen and Hao Zhang.
\newblock Learning implicit fields for generative shape modeling.
\newblock In {\em CVPR}, 2019.

\bibitem{chibane2020neural}
Julian Chibane, Aymen Mir, and Gerard Pons-Moll.
\newblock Neural unsigned distance fields for implicit function learning.
\newblock In {\em NeurIPS}, 2020.

\bibitem{chibane2020implicit}
Julian Chibane and Gerard Pons-Moll.
\newblock Implicit feature networks for texture completion from partial 3d
  data.
\newblock In {\em European Conference on Computer Vision}, pages 717--725.
  Springer, 2020.

\bibitem{dai2017scannet}
Angela Dai, Angel~X Chang, Manolis Savva, Maciej Halber, Thomas Funkhouser, and
  Matthias Nie{\ss}ner.
\newblock Scannet: Richly-annotated 3d reconstructions of indoor scenes.
\newblock In {\em Proceedings of the IEEE conference on computer vision and
  pattern recognition}, pages 5828--5839, 2017.

\bibitem{deng2020cvxnet}
Boyang Deng, Kyle Genova, Soroosh Yazdani, Sofien Bouaziz, Geoffrey Hinton, and
  Andrea Tagliasacchi.
\newblock Cvxnet: Learnable convex decomposition.
\newblock In {\em CVPR}, 2020.

\bibitem{deprelle2019learning}
Theo Deprelle, Thibault Groueix, Matthew Fisher, Vladimir~G Kim, Bryan~C
  Russell, and Mathieu Aubry.
\newblock Learning elementary structures for 3d shape generation and matching.
\newblock In {\em NeurIPS}, 2019.

\bibitem{erler2020points2surf}
Philipp Erler, Paul Guerrero, Stefan Ohrhallinger, Niloy~J Mitra, and Michael
  Wimmer.
\newblock Points2surf learning implicit surfaces from point clouds.
\newblock In {\em ECCV}, 2020.

\bibitem{fan2017point}
Haoqiang Fan, Hao Su, and Leonidas~J Guibas.
\newblock A point set generation network for 3d object reconstruction from a
  single image.
\newblock In {\em CVPR}, 2017.

\bibitem{genova2020local}
Kyle Genova, Forrester Cole, Avneesh Sud, Aaron Sarna, and Thomas Funkhouser.
\newblock Local deep implicit functions for 3d shape.
\newblock In {\em CVPR}, 2020.

\bibitem{genova2019learning}
Kyle Genova, Forrester Cole, Daniel Vlasic, Aaron Sarna, William~T Freeman, and
  Thomas Funkhouser.
\newblock Learning shape templates with structured implicit functions.
\newblock In {\em ICCV}, 2019.

\bibitem{gropp2020implicit}
Amos Gropp, Lior Yariv, Niv Haim, Matan Atzmon, and Yaron Lipman.
\newblock Implicit geometric regularization for learning shapes.
\newblock In {\em ICML}, 2020.

\bibitem{groueix2018papier}
Thibault Groueix, Matthew Fisher, Vladimir~G Kim, Bryan~C Russell, and Mathieu
  Aubry.
\newblock A papier-m{\^a}ch{\'e} approach to learning 3d surface generation.
\newblock In {\em CVPR}, 2018.

\bibitem{guennebaud2007algebraic}
Ga{\"e}l Guennebaud and Markus Gross.
\newblock Algebraic point set surfaces.
\newblock In {\em ACM siggraph 2007 papers}, pages 23--es. 2007.

\bibitem{hart1996sphere}
John~C Hart.
\newblock Sphere tracing: A geometric method for the antialiased ray tracing of
  implicit surfaces.
\newblock {\em The Visual Computer}, 1996.

\bibitem{huang2023neural}
Jiahui Huang, Zan Gojcic, Matan Atzmon, Or Litany, Sanja Fidler, and Francis
  Williams.
\newblock Neural kernel surface reconstruction.
\newblock In {\em Proceedings of the IEEE/CVF Conference on Computer Vision and
  Pattern Recognition}, pages 4369--4379, 2023.

\bibitem{jain2021dreamfields}
Ajay Jain, Ben Mildenhall, Jonathan~T. Barron, Pieter Abbeel, and Ben Poole.
\newblock Zero-shot text-guided object generation with dream fields.
\newblock 2022.

\bibitem{jiang2020local}
Chiyu Jiang, Avneesh Sud, Ameesh Makadia, Jingwei Huang, Matthias Nie{\ss}ner,
  Thomas Funkhouser, et~al.
\newblock Local implicit grid representations for 3d scenes.
\newblock In {\em CVPR}, 2020.

\bibitem{kato2018neural}
Hiroharu Kato, Yoshitaka Ushiku, and Tatsuya Harada.
\newblock Neural 3d mesh renderer.
\newblock In {\em CVPR}, 2018.

\bibitem{kazhdan2013screened}
Michael Kazhdan and Hugues Hoppe.
\newblock Screened poisson surface reconstruction.
\newblock {\em TOG}, 2013.

\bibitem{kolluri2008provably}
Ravikrishna Kolluri.
\newblock Provably good moving least squares.
\newblock {\em TALG}, 2008.

\bibitem{li2022learning}
Tianyang Li, Xin Wen, Yu-Shen Liu, Hua Su, and Zhizhong Han.
\newblock Learning deep implicit functions for 3d shapes with dynamic code
  clouds.
\newblock In {\em Proceedings of the IEEE/CVF Conference on Computer Vision and
  Pattern Recognition}, pages 12840--12850, 2022.

\bibitem{lin2022surface}
Siyou Lin, Dong Xiao, Zuoqiang Shi, and Bin Wang.
\newblock Surface reconstruction from point clouds without normals by
  parametrizing the gauss formula.
\newblock {\em ACM Transactions on Graphics}, 42(2):1--19, 2022.

\bibitem{lionar2021dynamic}
Stefan Lionar, Daniil Emtsev, Dusan Svilarkovic, and Songyou Peng.
\newblock Dynamic plane convolutional occupancy networks.
\newblock In {\em Proceedings of the IEEE/CVF Winter Conference on Applications
  of Computer Vision}, pages 1829--1838, 2021.

\bibitem{lipman2021phase}
Yaron Lipman.
\newblock Phase transitions, distance functions, and implicit neural
  representations.
\newblock In {\em ICML}, 2021.

\bibitem{liu2018planenet}
Chen Liu, Jimei Yang, Duygu Ceylan, Ersin Yumer, and Yasutaka Furukawa.
\newblock Planenet: Piece-wise planar reconstruction from a single rgb image.
\newblock In {\em CVPR}, 2018.

\bibitem{liu2022learning}
Hsueh-Ti~Derek Liu, Francis Williams, Alec Jacobson, Sanja Fidler, and Or
  Litany.
\newblock Learning smooth neural functions via lipschitz regularization.
\newblock {\em arXiv preprint arXiv:2202.08345}, 2022.

\bibitem{liu2020meshing}
Minghua Liu, Xiaoshuai Zhang, and Hao Su.
\newblock Meshing point clouds with predicted intrinsic-extrinsic ratio
  guidance.
\newblock In {\em ECCV}, 2020.

\bibitem{liu2021deep}
Shi-Lin Liu, Hao-Xiang Guo, Hao Pan, Peng-Shuai Wang, Xin Tong, and Yang Liu.
\newblock Deep implicit moving least-squares functions for 3d reconstruction.
\newblock In {\em CVPR}, 2021.

\bibitem{lorensen1987marching}
William~E Lorensen and Harvey~E Cline.
\newblock Marching cubes: A high resolution 3d surface construction algorithm.
\newblock In {\em SIGGRAPH}, 1987.

\bibitem{ma2020neural}
Baorui Ma, Zhizhong Han, Yu-Shen Liu, and Matthias Zwicker.
\newblock Neural-pull: Learning signed distance functions from point clouds by
  learning to pull space onto surfaces.
\newblock In {\em ICML}, 2021.

\bibitem{ma2022reconstructing}
Baorui Ma, Yu-Shen Liu, and Zhizhong Han.
\newblock Reconstructing surfaces for sparse point clouds with on-surface
  priors.
\newblock In {\em Proceedings of the IEEE/CVF Conference on Computer Vision and
  Pattern Recognition}, pages 6315--6325, 2022.

\bibitem{ma2022surface}
Baorui Ma, Yu-Shen Liu, Matthias Zwicker, and Zhizhong Han.
\newblock Surface reconstruction from point clouds by learning predictive
  context priors.
\newblock In {\em Proceedings of the IEEE/CVF Conference on Computer Vision and
  Pattern Recognition}, pages 6326--6337, 2022.

\bibitem{meanti2022efficient}
Giacomo Meanti, Luigi Carratino, Ernesto De~Vito, and Lorenzo Rosasco.
\newblock Efficient hyperparameter tuning for large scale kernel ridge
  regression.
\newblock In {\em International Conference on Artificial Intelligence and
  Statistics}, pages 6554--6572. PMLR, 2022.

\bibitem{meanti2020kernel}
Giacomo Meanti, Luigi Carratino, Lorenzo Rosasco, and Alessandro Rudi.
\newblock Kernel methods through the roof: handling billions of points
  efficiently.
\newblock {\em Advances in Neural Information Processing Systems},
  33:14410--14422, 2020.

\bibitem{mercier2022moving}
Corentin Mercier, Thibault Lescoat, Pierre Roussillon, Tamy Boubekeur, and
  Jean-Marc Thiery.
\newblock Moving level-of-detail surfaces.
\newblock {\em ACM Transactions on Graphics (TOG)}, 41(4):1--10, 2022.

\bibitem{mescheder2019occupancy}
Lars Mescheder, Michael Oechsle, Michael Niemeyer, Sebastian Nowozin, and
  Andreas Geiger.
\newblock Occupancy networks: Learning 3d reconstruction in function space.
\newblock In {\em Proceedings of the IEEE/CVF conference on computer vision and
  pattern recognition}, pages 4460--4470, 2019.

\bibitem{mildenhall2020nerf}
Ben Mildenhall, Pratul~P Srinivasan, Matthew Tancik, Jonathan~T Barron, Ravi
  Ramamoorthi, and Ren Ng.
\newblock Nerf: Representing scenes as neural radiance fields for view
  synthesis.
\newblock In {\em ECCV}, 2020.

\bibitem{ouasfi2022few}
Amine Ouasfi and Adnane Boukhayma.
\newblock Few'zero level set'-shot learning of shape signed distance functions
  in feature space.
\newblock In {\em ECCV}, 2022.

\bibitem{ouasfi2024mix}
Amine Ouasfi and Adnane Boukhayma.
\newblock Mixing-denoising generalizable occupancy networks.
\newblock In {\em 3DV}, 2024.

\bibitem{palmer2022deepcurrents}
David Palmer, Dmitriy Smirnov, Stephanie Wang, Albert Chern, and Justin
  Solomon.
\newblock Deepcurrents: Learning implicit representations of shapes with
  boundaries.
\newblock In {\em Proceedings of the IEEE/CVF Conference on Computer Vision and
  Pattern Recognition}, pages 18665--18675, 2022.

\bibitem{park2019deepsdf}
Jeong~Joon Park, Peter Florence, Julian Straub, Richard Newcombe, and Steven
  Lovegrove.
\newblock Deepsdf: Learning continuous signed distance functions for shape
  representation.
\newblock In {\em CVPR}, 2019.

\bibitem{peng2021shape}
Songyou Peng, Chiyu Jiang, Yiyi Liao, Michael Niemeyer, Marc Pollefeys, and
  Andreas Geiger.
\newblock Shape as points: A differentiable poisson solver.
\newblock {\em Advances in Neural Information Processing Systems},
  34:13032--13044, 2021.

\bibitem{peng2020convolutional}
Songyou Peng, Michael Niemeyer, Lars Mescheder, Marc Pollefeys, and Andreas
  Geiger.
\newblock Convolutional occupancy networks.
\newblock In {\em European Conference on Computer Vision}, pages 523--540.
  Springer, 2020.

\bibitem{qi2017pointnet}
Charles~R Qi, Hao Su, Kaichun Mo, and Leonidas~J Guibas.
\newblock Pointnet: Deep learning on point sets for 3d classification and
  segmentation.
\newblock In {\em CVPR}, 2017.

\bibitem{rakotosaona2021differentiable}
Marie-Julie Rakotosaona, Noam Aigerman, Niloy Mitra, Maks Ovsjanikov, and Paul
  Guerrero.
\newblock Differentiable surface triangulation.
\newblock In {\em SIGGRAPH Asia}, 2021.

\bibitem{riegler2017octnet}
Gernot Riegler, Ali Osman~Ulusoy, and Andreas Geiger.
\newblock Octnet: Learning deep 3d representations at high resolutions.
\newblock In {\em CVPR}, 2017.

\bibitem{scholkopf2004kernel}
Bernhard Sch{\"o}lkopf, Joachim Giesen, and Simon Spalinger.
\newblock Kernel methods for implicit surface modeling.
\newblock In {\em NeurIPS}, 2004.

\bibitem{schoenberger2016sfm}
Johannes~Lutz Sch\"{o}nberger and Jan-Michael Frahm.
\newblock Structure-from-motion revisited.
\newblock In {\em Conference on Computer Vision and Pattern Recognition
  (CVPR)}, 2016.

\bibitem{schoenberger2016mvs}
Johannes~Lutz Sch\"{o}nberger, Enliang Zheng, Marc Pollefeys, and Jan-Michael
  Frahm.
\newblock Pixelwise view selection for unstructured multi-view stereo.
\newblock In {\em European Conference on Computer Vision (ECCV)}, 2016.

\bibitem{sitzmann2020metasdf}
Vincent Sitzmann, Eric~R Chan, Richard Tucker, Noah Snavely, and Gordon
  Wetzstein.
\newblock Metasdf: Meta-learning signed distance functions.
\newblock In {\em NeurIPS}, 2020.

\bibitem{sitzmann2020implicit}
Vincent Sitzmann, Julien Martel, Alexander Bergman, David Lindell, and Gordon
  Wetzstein.
\newblock Implicit neural representations with periodic activation functions.
\newblock In {\em NeurIPS}, 2020.

\bibitem{sitzmann2021light}
Vincent Sitzmann, Semon Rezchikov, William~T Freeman, Joshua~B Tenenbaum, and
  Fredo Durand.
\newblock Light field networks: Neural scene representations with
  single-evaluation rendering.
\newblock In {\em NeurIPS}, 2021.

\bibitem{NEURIPS2019_b5dc4e5d}
Vincent Sitzmann, Michael Zollhoefer, and Gordon Wetzstein.
\newblock Scene representation networks: Continuous 3d-structure-aware neural
  scene representations.
\newblock In {\em NeurIPS}, 2019.

\bibitem{takikawa2021neural}
Towaki Takikawa, Joey Litalien, Kangxue Yin, Karsten Kreis, Charles Loop, Derek
  Nowrouzezahrai, Alec Jacobson, Morgan McGuire, and Sanja Fidler.
\newblock Neural geometric level of detail: Real-time rendering with implicit
  3d shapes.
\newblock In {\em CVPR}, 2021.

\bibitem{tang2021sa}
Jiapeng Tang, Jiabao Lei, Dan Xu, Feiying Ma, Kui Jia, and Lei Zhang.
\newblock Sa-convonet: Sign-agnostic optimization of convolutional occupancy
  networks.
\newblock In {\em Proceedings of the IEEE/CVF International Conference on
  Computer Vision}, pages 6504--6513, 2021.

\bibitem{tatarchenko2017octree}
Maxim Tatarchenko, Alexey Dosovitskiy, and Thomas Brox.
\newblock Octree generating networks: Efficient convolutional architectures for
  high-resolution 3d outputs.
\newblock In {\em ICCV}, 2017.

\bibitem{tretschk2020patchnets}
Edgar Tretschk, Ayush Tewari, Vladislav Golyanik, Michael Zollh{\"o}fer,
  Carsten Stoll, and Christian Theobalt.
\newblock Patchnets: Patch-based generalizable deep implicit 3d shape
  representations.
\newblock In {\em ECCV}, 2020.

\bibitem{abstractionTulsiani17}
Shubham Tulsiani, Hao Su, Leonidas~J. Guibas, Alexei~A. Efros, and Jitendra
  Malik.
\newblock Learning shape abstractions by assembling volumetric primitives.
\newblock In {\em CVPR}, 2017.

\bibitem{wang2018pixel2mesh}
Nanyang Wang, Yinda Zhang, Zhuwen Li, Yanwei Fu, Wei Liu, and Yu-Gang Jiang.
\newblock Pixel2mesh: Generating 3d mesh models from single rgb images.
\newblock In {\em ECCV}, 2018.

\bibitem{wang2021neus}
Peng Wang, Lingjie Liu, Yuan Liu, Christian Theobalt, Taku Komura, and Wenping
  Wang.
\newblock Neus: Learning neural implicit surfaces by volume rendering for
  multi-view reconstruction.
\newblock {\em arXiv preprint arXiv:2106.10689}, 2021.

\bibitem{wang2017cnn}
Peng-Shuai Wang, Yang Liu, Yu-Xiao Guo, Chun-Yu Sun, and Xin Tong.
\newblock O-cnn: Octree-based convolutional neural networks for 3d shape
  analysis.
\newblock {\em TOG}, 2017.

\bibitem{wang2021metaavatar}
Shaofei Wang, Marko Mihajlovic, Qianli Ma, Andreas Geiger, and Siyu Tang.
\newblock Metaavatar: Learning animatable clothed human models from few depth
  images.
\newblock {\em Advances in Neural Information Processing Systems},
  34:2810--2822, 2021.

\bibitem{williams2000using}
Christopher Williams and Matthias Seeger.
\newblock Using the nystr{\"o}m method to speed up kernel machines.
\newblock {\em Advances in neural information processing systems}, 13, 2000.

\bibitem{williams2022neural}
Francis Williams, Zan Gojcic, Sameh Khamis, Denis Zorin, Joan Bruna, Sanja
  Fidler, and Or Litany.
\newblock Neural fields as learnable kernels for 3d reconstruction.
\newblock In {\em CVPR}, 2022.

\bibitem{williams2019deep}
Francis Williams, Teseo Schneider, Claudio Silva, Denis Zorin, Joan Bruna, and
  Daniele Panozzo.
\newblock Deep geometric prior for surface reconstruction.
\newblock In {\em CVPR}, 2019.

\bibitem{williams2021neural}
Francis Williams, Matthew Trager, Joan Bruna, and Denis Zorin.
\newblock Neural splines: Fitting 3d surfaces with infinitely-wide neural
  networks.
\newblock In {\em CVPR}, 2021.

\bibitem{wu2016learning}
Jiajun Wu, Chengkai Zhang, Tianfan Xue, William~T Freeman, and Joshua~B
  Tenenbaum.
\newblock Learning a probabilistic latent space of object shapes via 3d
  generative-adversarial modeling.
\newblock In {\em NeurIPS}, 2016.

\bibitem{wu20153d}
Zhirong Wu, Shuran Song, Aditya Khosla, Fisher Yu, Linguang Zhang, Xiaoou Tang,
  and Jianxiong Xiao.
\newblock 3d shapenets: A deep representation for volumetric shapes.
\newblock In {\em CVPR}, 2015.

\bibitem{yariv2021volume}
Lior Yariv, Jiatao Gu, Yoni Kasten, and Yaron Lipman.
\newblock Volume rendering of neural implicit surfaces.
\newblock {\em Advances in Neural Information Processing Systems},
  34:4805--4815, 2021.

\bibitem{yavartanoo20213dias}
Mohsen Yavartanoo, Jaeyoung Chung, Reyhaneh Neshatavar, and Kyoung~Mu Lee.
\newblock 3dias: 3d shape reconstruction with implicit algebraic surfaces.
\newblock In {\em ICCV}, 2021.

\bibitem{Zhou2022CAP-UDF}
Junsheng Zhou, Baorui Ma, Liu Yu-Shen, Fang Yi, and Han Zhizhong.
\newblock Learning consistency-aware unsigned distance functions progressively
  from raw point clouds.
\newblock In {\em Advances in Neural Information Processing Systems (NeurIPS)},
  2022.

\bibitem{zou20173d}
Chuhang Zou, Ersin Yumer, Jimei Yang, Duygu Ceylan, and Derek Hoiem.
\newblock 3d-prnn: Generating shape primitives with recurrent neural networks.
\newblock In {\em CVPR}, 2017.

\end{thebibliography}
}

\clearpage
\section{Supplementary Material}

\subsection{Quantitative comparison to SPSR \cite{kazhdan2013screened} in ShapeNet \cite{shapenet}}  

\begin{table}[h!]
\centering
\caption{Comparison to PCA normals +  SPSR.}
\scalebox{0.9}{
\begin{tabular}{@{}lcccc@{}}
        \toprule
        {} & IoU $\uparrow$ & CD$_1${$\downarrow$} & NC$\uparrow$ & FS$\uparrow$ \\
        \midrule
        SPSR & 0.65 & 1.28 & 0.84 & 0.74\\
        Ours & \textbf{0.90} & \textbf{0.35} & \textbf{0.95} & \textbf{0.98}\\
        \bottomrule
        \label{tab:spsr}
\end{tabular}}
\end{table}  

Table \ref{tab:spsr} shows a numerical comparison to Screened Poisson Surface Reconstruction (SPSR) \cite{kazhdan2013screened} on classes Tables, Lamps and Chairs of ShapeNet \cite{shapenet}. Input point clouds are of size 10k with 0.005 standard deviation Gaussian noise. We remind that our method builds on Poco \cite{boulch2022poco}.   

\subsection{Additional Qualitative Results for ShapeNet \cite{shapenet} Reconstructions}

\begin{figure}[h!]
\centering
\includegraphics[width=\linewidth]{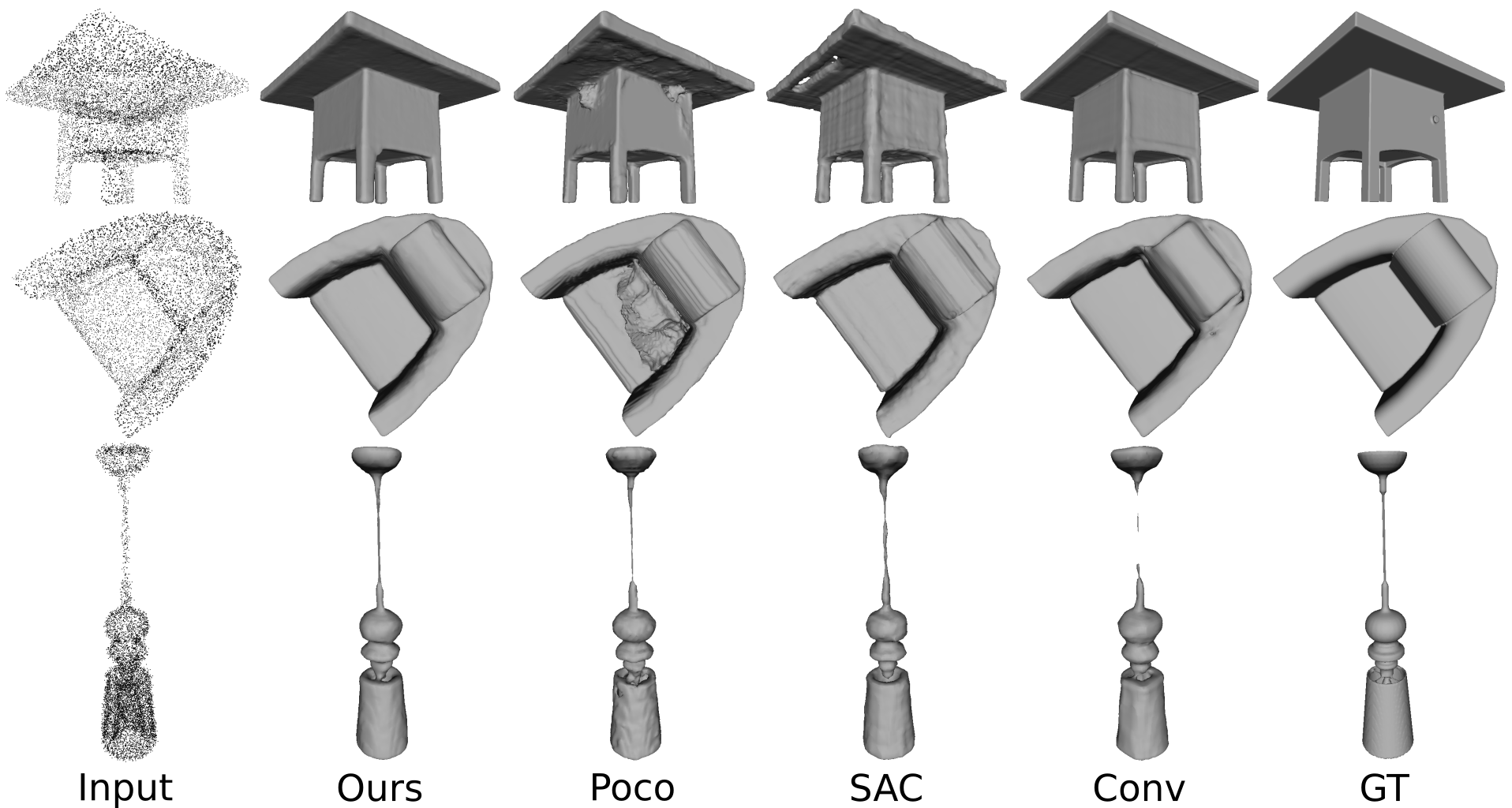}
  \caption{Additional qualitative comparison of ShapeNet reconstructions.}
  \label{fig:sn}
\end{figure}

Figure \ref{fig:sn} shows additional reconstruction examples from our method, Poco \cite{boulch2022poco}, SAC \cite{tang2021sa} and Conv \cite{peng2020convolutional}. Input point clouds are of size 10k with 0.005 standard deviation Gaussian noise. 
While test-time finetuning (SAC) can improve on its baseline (Conv), it can also lead to worse final reconstructions. Notice how our approach can reliably improve on its beseline (Poco).  

\subsection{Recovering from baseline failures}

\begin{figure}[h!]
\centering
\includegraphics[width=.6\linewidth]{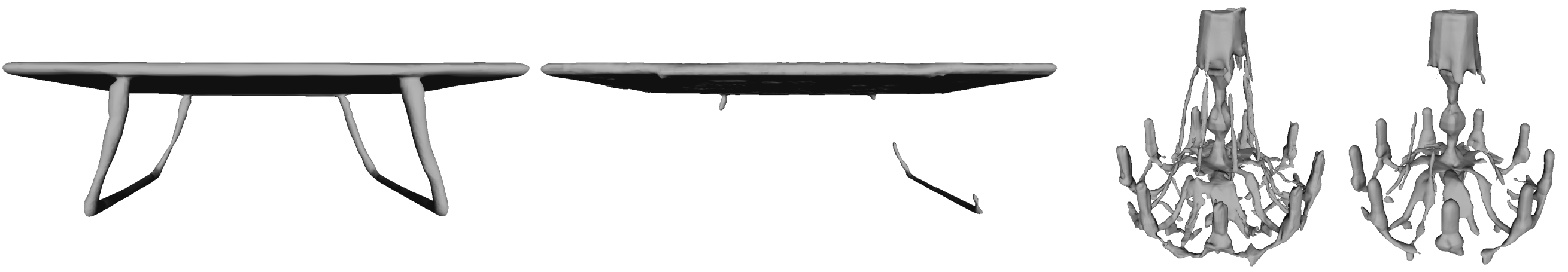}
  \caption{Comparison to our baseline in ShapeNet \cite{shapenet}. \textbf{Ours / Baseline}.}
\label{fig:fail}  
\end{figure}

Figure \ref{fig:fail} shows cases where we can recover from failures of our baseline (Poco \cite{boulch2022poco}). Input point clouds are of size 10k plus a 0.005 standard deviation Gaussian noise.  We improve \textbf{chamfer error} from \textbf{0.85 (baseline)} to \textbf{0.27 (Ours)} for the table example on the left, and from \textbf{1.22 (baseline)} to \textbf{0.86 (Ours)} for the chandelier on the right. As part of our limitations, and despite improving on the reconstruction of the network baseline, it is still challenging for our method to recover the finest structures and intricate details of a complex shape such as the chandelier example. 

%\section*{Qualitative Comparison to NP \cite{ma2020neural}}
\subsection{Evaluation Metrics}

Following the definitions from \cite{boulch2022poco}, we present here the formal definitions for the metrics that we use for evaluation in the main submission. We denote by $\mathcal{S}$ and  $\hat{\mathcal{S}}$ the ground truth and predicted mesh respectively. All metrics are approximated with 100k samples from $\mathcal{S}$ and  $\hat{\mathcal{S}}$. 

\paragraph{Intersection over Union (IoU)}
The volumetric Intersection over Union is the ratio of  the intersection of the inside volumes of the meshes to their union:
$$
\text { IoU }=\frac{\mid\{x \in \Omega: x \text { inside } \mathcal{S} \text { and } \hat{\mathcal{S}}\} \mid}{\mid\{x \in \Omega: x \text { inside } \mathcal{S} \text { or } \hat{\mathcal{S}}\} \mid}
,$$
where $ |.| $ represents the sets cardinality, which is approximately determined by sampling $100 \mathrm{k}$ points from the bounding volume $\Omega$ of $\mathcal{S}$. 

\paragraph{Chamfer Distance (CD$_1$)} The L$_1$ Chamfer distance is based on the two-ways nearest neighbor distance: 
$$\mathrm{CD}_1=\frac{1}{2|\mathcal{S}|} \sum_{v \in \mathcal{S}} \min _{\hat{v} \in \hat{\mathcal{S}}}\|v-\hat{v}\|_2+\frac{1}{2|\hat{\mathcal{S}}|} \sum_{\hat{v} \in \hat{\mathcal{S}}} \min _{v \in \mathcal{S}}\|\hat{v}-v\|_2.$$

\paragraph{F-Score (FS)} For a given threshold $\tau$, the F-score between  the meshes $\mathcal{S}$ and $\hat{\mathcal{S}}$ is defined as:
$$
\mathrm{FS}\left(\tau, \mathcal{S}, \hat{\mathcal{S}}\right)=\frac{2 \text { Recall} \cdot \text{Precision }}{\text { Recall }+\text { Precision }},
$$

where
$$
\begin{array}{r}
\operatorname{Recall}\left(\tau, \mathcal{S}, \hat{\mathcal{S}}\right)=\mid\left\{v \in \mathcal{S} \text {, s.t. } \min _{\hat{v} \in \hat{ \mathcal{S} }} \left\|v-\hat{v}\|_2\right<\tau\right\} \mid ,\\
\operatorname{Precision}\left(\tau, \mathcal{S}, \hat{\mathcal{S}}\right)=\mid\left\{\hat{v} \in \hat{\mathcal{S} }\text {, s.t. } \min _{v \in  \mathcal{S} } \left\|v-\hat{v}\|_2\right<\tau\right\} \mid .\\
\end{array}
$$
Following \cite{mescheder2019occupancy} and \cite{peng2020convolutional}, we set $\tau$ to $0.01$.

\paragraph{Normal consistency (NC)} We denote here by $n_v$ the normal at a point $v$ in $\mathcal{S}$. The normal consistency between two meshes $\mathcal{S}$ and $\hat{\mathcal{S}}$ is defined as: 

$$\mathrm{NC}=\frac{1}{2|\mathcal{S}|} \sum_{v \in \mathcal{S}} n_{v} \cdot n_{\operatorname{closest}(v,\hat{ \mathcal{S}})}+\frac{1}{2|\hat{\mathcal{S}}|} \sum_{\hat{v} \in \hat{\mathcal{S}}} n_{\hat{v}} \cdot n_{\operatorname{closest}(\hat{v}, \mathcal{S})},$$

where  
$$
\operatorname{closest}(v, \hat{\mathcal{S}}) = \operatorname{argmin} _{\hat{v} \in \hat{\mathcal{S}}}\|v-\hat{v}\|_2.
$$

%%%%%%%%%%%%%%%%%%%%%%%%%%%%%%%%%%%%%%%%%%%%%%%%%%%%%%%%%%%%
\end{document}